\documentclass[conference]{IEEEtran}
\IEEEoverridecommandlockouts
\usepackage{cite}
\usepackage{amsmath,amssymb,amsfonts}
\usepackage{graphicx}
\usepackage{textcomp}
\usepackage{xcolor}
\usepackage{fancyhdr}
\usepackage[hyphens]{url}
\usepackage{hyperref}
\usepackage{algorithm}
\usepackage{algpseudocode}
\usepackage[T1]{fontenc}
\usepackage[utf8]{inputenc}
\usepackage{booktabs}
\usepackage{multirow}
\usepackage{algorithm}
\usepackage{algpseudocode}
\usepackage{tikz}
\usepackage{xspace}
\usepackage{array}
\usepackage[normalem]{ulem}
\usepackage{etoolbox}
\usepackage{tikz}
\usepackage{placeins}
\usepackage{subfig}
\usepackage{soul}
\def\BibTeX{{\rm B\kern-.05em{\sc i\kern-.025em b}\kern-.08em
    T\kern-.1667em\lower.7ex\hbox{E}\kern-.125emX}}
\begin{document}

\newcommand{\proposed}{\textsc{Pascal}\xspace}
\newcommand{\nomigration}{\textsc{Pascal}(NoMigration)\xspace}
\newcommand{\nonadaptive}{\textsc{Pascal}(NonAdaptive)\xspace}

\title{\proposed: A Phase-Aware Scheduling Algorithm for Serving Reasoning-based Large Language Models\\
}
\author{
\IEEEauthorblockN{Eunyeong Cho}
\IEEEauthorblockA{
    {KAIST} \\
    eunyeong.cho@kaist.ac.kr
}
\and
\IEEEauthorblockN{Jehyeon Bang}
\IEEEauthorblockA{
    {KAIST} \\
    jehyeon.bang@kaist.ac.kr
}
\and
\IEEEauthorblockN{Ranggi Hwang\IEEEauthorrefmark{1}}
\IEEEauthorblockA{
    {UNIST} \\
    ranggi.hwang@unist.ac.kr
}
\and
\IEEEauthorblockN{Minsoo Rhu}
\IEEEauthorblockA{
    {KAIST} \\
    mrhu@kaist.ac.kr
}
\thanks{\protect{\IEEEauthorrefmark{1}Work done while at KAIST.}}
}

\maketitle

\newcommand{\hpcayear}{2026}

\newcommand{\circlednum}[1]{\ignorespacesafterend
  \tikz[baseline=-0.8ex] \node[draw, circle, fill=black, text=white, inner sep=0.05mm]{#1};\ignorespacesafterend
}
\robustify{\circlednum}

\newcommand{\wcirclednum}[1]{\ignorespacesafterend
  \tikz[baseline=-0.8ex] \node[draw, circle, fill=white, text=black, inner sep=0.05mm]{#1};\ignorespacesafterend
}
\robustify{\wcirclednum}

\newcommand{\fig}[1]{Figure~\ref{#1}}
\newcommand{\equat}[1]{Equation~\ref{#1}}
\newcommand{\sect}[1]{Section~\ref{#1}}
\newcommand{\tab}[1]{Table~\ref{#1}}
\newcommand{\algo}[1]{Algorithm~\ref{#1}}
\newcommand{\eqn}[1]{Equation~\ref{#1}}
\newcommand{\circled}[1]{\tikz[baseline=(char.base)]{
  \node[shape=circle,draw,inner sep=1pt] (char) {#1};}}

\title{\proposed: A Phase-Aware Scheduling Algorithm for Serving Reasoning-based Large Language Models}

\begin{abstract}

The emergence of reasoning-based LLMs leveraging Chain-of-Thought (CoT) inference introduces new serving challenges, as their extended reasoning phases delay user-visible output and inflate Time-To-First-Token (TTFT). Existing LLM serving frameworks fail to distinguish between reasoning and answering phases, leading to performance degradation under GPU memory constraints. We present \proposed, a phase-aware scheduling algorithm that prioritizes reasoning to reduce TTFT while using controlled preemption and token pacing during answering to preserve Quality-of-Experience (QoE). Our hierarchical scheduler combines instance-level placement with intra-instance execution and enables dynamic migration at phase boundaries to balance load and reduce interference. Across benchmarks using DeepSeek-R1-Distill-Qwen-32B, \proposed reduces tail TTFT by up to 72\% while maintaining answering phase SLO attainment, demonstrating the importance of phase-aware scheduling for reasoning-based LLM deployment.

\end{abstract}

\section{Introduction}
\label{sect:introduction}

Large language models (LLMs)~\cite{gpt4, llama3, palm, gemini} have emerged as powerful tools for various natural language processing tasks. As these models evolve, their  deployment for real-time applications has become increasingly important, with a particular focus on inference efficiency and user experience. In LLM inference, user experience is significantly influenced by two key performance metrics. \emph{Time-To-First-Token} (TTFT) measures the latency between the query submission time and the time the first response token is received. \emph{Time-Per-Output-Token} (TPOT), on the other hand, measures the generation speed of each subsequent token. Prior work has established that these metrics correlate strongly with user satisfaction and engagement~\cite{deepspeed_fastgen, andes, distserve, throttling, aqua, qlm, sarathi}, making them critical design objectives when LLMs are deployed.

\begin{figure}
    \centering
    \includegraphics[width=0.96\columnwidth]{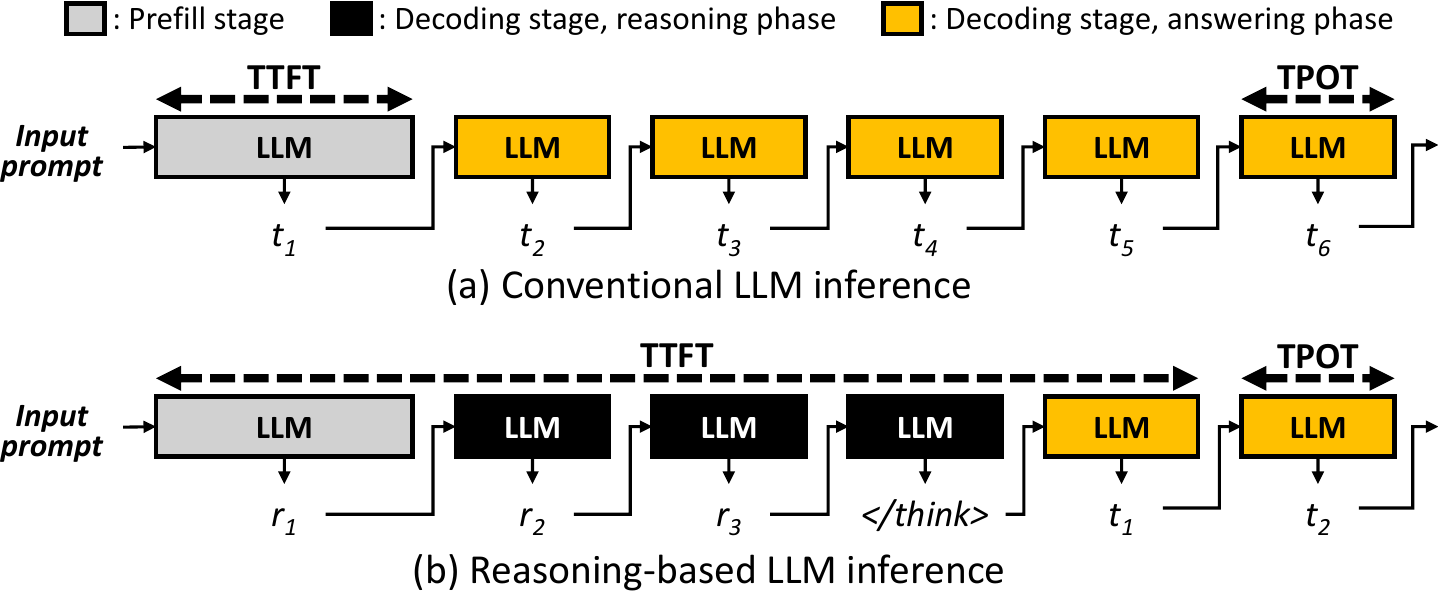}
    \caption{
Comparison of how TTFT and TPOT are defined in conventional LLMs versus reasoning-based LLMs. {TTFT is the latency between the query submission time and the time the first ``user-visible'' response token ($t_1$) is received by the user. (a) In conventional LLMs, this first user-visible token is generated at the end of the prefill stage, so TTFT equals the prefill-stage latency. (b) In contrast, for reasoning-based LLMs, TTFT is the sum of the prefill-stage latency, the latency of the reasoning phase, and the latency from the end of the reasoning phase to the generation of the first answering token.}
    }\label{fig:ttft_and_tpot}
    \vspace{-1.3em}
\end{figure}

These user experience metrics are directly tied to the underlying computational stages of LLM inference. As shown in \fig{fig:ttft_and_tpot}(a), conventional LLM inference consists of two distinct stages: (1) the \emph{prefill} stage processes the entire input prompt in a single forward pass and primarily determines TTFT, while (2) the \emph{decoding} stage generates tokens auto-regressively, directly influencing TPOT. This clear distinction between stages has enabled prior works to develop targeted optimization strategies that align the prefill stage and  decoding stage with user satisfaction metrics. Moreover, recent serving systems leverage the distinct computational characteristics of the prefill and decoding stages to better meet \emph{service-level objectives (SLOs)} for TTFT and TPOT {through disaggregated execution of prefill and decoding stage.}~\cite{splitwise, distserve, dynamo, dynamollm}.

Although these approaches have been effective for conventional LLMs, the research landscape has evolved significantly with the emergence of \emph{reasoning}-based LLMs~\cite{deepseekr1, o4-mini, qwq32b, reasoning_model1}. Reasoning models incorporate Chain-of-Thought (CoT) techniques~\cite{cot_prompt, cot_tree, self_cot, least_cot, other_cot1, other_cot2}, which have gained tremendous popularity for their ability to solve complex problems by explicitly generating \emph{intermediate} reasoning steps ``before'' producing final answers. We observe that this fundamental shift in how models generate outputs introduces new challenges for inference optimization, as detailed below.

Unlike conventional LLMs, reasoning-based LLMs produce intermediate \emph{reasoning tokens} that represent step-by-step problem-solving, followed by \emph{answering tokens} that convey the final outcome (\fig{fig:ttft_and_tpot}(b)).
This distinction disrupts the conventional relationship between performance metrics (i.e., TTFT and TPOT) and the serving pipeline stages (i.e., prefill and decoding stages).
Specifically, in reasoning-based LLMs, the decoding stage itself is divided into two functionally distinct phases: a \emph{reasoning phase}, where reasoning tokens are generated, and an \emph{answering phase} that produces the final answering tokens.
The key insight here is that {the performance and SLO requirements} differ between these two phases. Because the user does not require direct insight into the model’s internal reasoning tokens, the top priority is to expedite {the reasoning phase} as fast as possible, reducing the time it takes to generate the model’s reasoning tokens before the final answering tokens start streaming in. By minimizing the reasoning phase latency, the serving system can begin transmitting the user-visible tokens (i.e., the answering tokens) earlier, improving the user-perceived responsiveness by minimizing the latency to generate the first meaningful answering token.
In contrast, the answering phase just has to be ``\emph{fast enough}'' to ensure a smooth and prompt user experience, as it can tolerate somewhat looser throughput targets compared to the user-hidden reasoning phase.
In other words, once the user begins seeing the answering tokens, maintaining a modestly high streaming rate is often sufficient for good user experience.
Overall, TTFT in a reasoning model effectively encompasses not only the prefill stage but also the decoding stage of potentially numerous intermediate reasoning tokens (which are not necessary for the user to comprehend the final answer) and the latency from the end of the reasoning phase to the generation of the first answering token (\fig{fig:ttft_and_tpot}(b)).

To this end, we present \proposed, a \underline{p}hase-\underline{a}ware \underline{sc}heduling \underline{al}gorithm that optimizes the distinct requirements of the reasoning and answering phases to enhance {user experience while maintaining serving throughput}. 
Our characterization of reasoning-based LLMs reveals the following key observations where (1) the absolute latency of the decoding stage, which directly impacts the reasoning model's TTFT, is highly sensitive to interruptions in the execution pipeline, while (2) its TPOT requirements can be maintained much more robustly with greater scheduling flexibility.
{Building on this insight, \proposed introduces a phase-aware, hierarchical scheduling architecture. Our proposal explicitly partitions the decoding stage into reasoning and answering phases, applies distinct scheduling priorities to each, and enables instance-level load balancing to address the unique challenges of serving reasoning models. Through extensive evaluation on state-of-the-art reasoning-based LLMs, we show that \proposed not only reduces the time to first answering token but also preserves or improves SLO satisfaction during the answering phase.}

\begin{figure*}[t!]
    \centering    \includegraphics[width=0.98\textwidth]{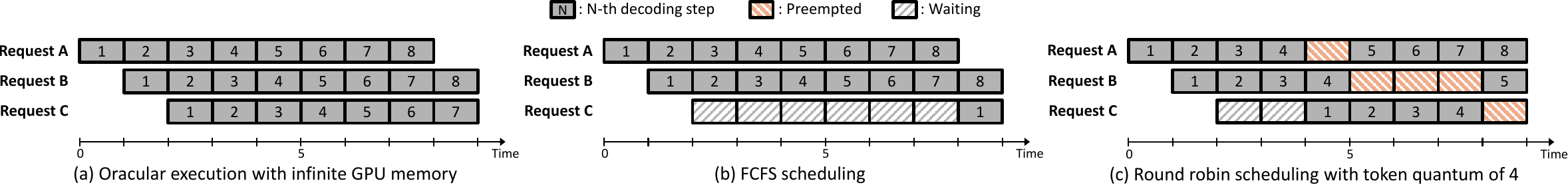}
    \caption{
Timeline illustrating how the serving system handles requests under three different scenarios. We assume that Request A, B, and C arrive at times 0, 1, and 2, respectively, and that the GPU memory capacity allows at most two requests to be batched simultaneously. (a) With oracular execution under infinite GPU memory, there is no limit on how many requests can be batched, so each request begins execution immediately upon arrival, and no preemption occurs.  (b) Under FCFS scheduling, the first two requests (A and B) are batched, while C must wait. Request C joins the batch only after A finishes decoding, yielding a TTFT of 7 time units. (c) In round‑robin (RR) scheduling, each request decodes a fixed quantum of four tokens before being preempted. Request C waits for 2 time units, joins the batch when Request A is preempted at time 4, and is itself preempted after producing four tokens at time 8.
}\label{fig:background_scheduling}
\end{figure*}

\section{Background}
\label{sect:background}

\subsection{LLM Serving Metrics and Execution Phases}
\label{sect:llm_serving}

LLM serving systems are commonly evaluated using two critical performance metrics: \emph{Time-To-First-Token} (TTFT), which measures how quickly a user receives the first generated token, and \emph{Time-Per-Output-Token} (TPOT), which quantifies the generation speed once decoding begins. These metrics correspond to distinct points in the response timeline, aligning with the two computational stages of LLM inference: the prefill and decoding stages (\fig{fig:ttft_and_tpot}(a)).

During the prefill stage, the model processes the entire input prompt in a single pass, generating key-value (KV) representations (referred to as KV caches) for all tokens, {and generating one output token (e.g., $t_1$ in \fig{fig:ttft_and_tpot}(a))}. This stage is compute-intensive, exhibits high parallelism, and directly determines TTFT in conventional LLMs. The subsequent decoding stage auto-regressively generates output tokens by reusing the KV caches and incrementally updating them with each newly generated token. This stage is known to be memory-bandwidth-bound and determines TPOT. In practice, LLM serving systems define a \emph{service-level objective (SLO)} to ensure a responsive user experience, typically targeting a TTFT of a few seconds or less and a TPOT of 5–10 tokens per second to align with human reading speeds~\cite{andes,distserve,deepspeed_fastgen}.

\subsection{Need for Blocking \& Preempting Requests in LLM Serving}

Modern LLM serving systems employ \emph{continuous batching}~\cite{orca}, where new inference requests are incrementally added to the currently running batch at each decoding step. During decoding, however, the KV cache grows as new tokens are generated. Given limited GPU memory, the number of concurrent requests that can be batched is therefore constrained by the cumulative KV cache footprint~\cite{flexgen, llminflash, alisa, h2o, pyramidkv}. Consequently, when GPU memory becomes saturated, the serving system must either \emph{block} newly arriving requests or \emph{preempt} (evict) in‑flight requests to free GPU memory. Concretely, blocking occurs when an incoming request cannot be scheduled due to insufficient memory; it waits in the queue until enough KV cache space is released by completing requests. Because execution cannot begin until admission, this queuing delay inflates the request’s TTFT. Preemption, in contrast, temporarily evicts an executing request; its KV cache is offloaded to CPU memory and execution is suspended. The request later resumes once sufficient GPU memory becomes available, reloading its KV cache to the GPU. Since preemption halts token generation, it increases TPOT.

\subsection{LLM Scheduling Policy}

An LLM serving system's scheduling policy determines which requests’ KV caches can remain resident in GPU memory and which should be blocked or preempted (and thus offloaded) to preserve GPU memory space. Early systems such as {\tt vLLM}\cite{vllm} rely on a simple \emph{First-Come-First-Served} (FCFS) scheduling policy that batches requests in arrival order. When GPU memory is exhausted under FCFS, the most recently arrived requests are preempted by offloading their KV caches to CPU memory. FCFS then blocks incoming requests until GPU memory becomes available, and preempted requests are resumed after their KV caches are reloaded from CPU to GPU memory. This mechanism significantly impacts TTFT, as new requests must wait for memory to free up, suffering from Head-of-Line (HoL) blocking where earlier long-running requests delay the servicing of later ones. \fig{fig:background_scheduling}(a) illustrates an ideal scenario without GPU memory constraints, where Request C can be immediately batched with earlier requests. In contrast, \fig{fig:background_scheduling}(b) shows a case where Request C must wait for Requests A and B to complete, resulting in an excessive TTFT of 7 time units. This is particularly problematic, as modern LLM serving systems define SLOs for both TTFT and TPOT to ensure responsiveness. FCFS inherently increases the risk of violating TTFT objectives due to HoL blocking under GPU memory pressure.

To address this limitation, recent work proposes various SLO-aware schedulers~\cite{aqua,fastserve,andes,pastfuture}. One line of this research explores \emph{time-sharing} within a single serving instance. In this context, a serving instance is the software abstraction in LLM serving frameworks that serves as the unit of execution and manages a replica of the model weights, coordinating access to shared GPU resources. Time-sharing schedulers~\cite{aqua,fastserve,andes} utilize preemption to allocate GPU time slices to concurrent requests, preventing any single long-running request from monopolizing the GPU memory. In addition to mitigating HoL blocking and improving TTFT, these approaches also help maintain acceptable TPOT by enabling fair resource allocation during decoding. A canonical example of time-sharing scheduler is the classical \emph{round-robin} (RR) algorithm. In this approach, the scheduler assigns each request a fixed \emph{token quantum}. Once a request consumes all its assigned quantum, its scheduling priority is lowered. If the GPU memory becomes limited, the scheduler can preempt the lowest-priority requests by offloading their KV caches to CPU memory. This frees up memory, allowing newly arrived requests to be admitted into the batch and begin decoding, thereby reducing their TTFT. More advanced schedulers are based on the same key insight, with priorities determined by different heuristics, for example, the likelihood of SLO violation~\cite{andes}, the number of generated tokens per request~\cite{aqua}, or estimated completion deadlines~\cite{fastserve}. As shown in \fig{fig:background_scheduling}(c), the earliest request, Request A, is preempted by the scheduler after generating a predefined token quantum (4 tokens in this example). This frees sufficient memory for Request C to join the batch and generate its first token within 3 time units.

\begin{figure}[t]
    \centering
    \includegraphics[width=0.94\columnwidth]{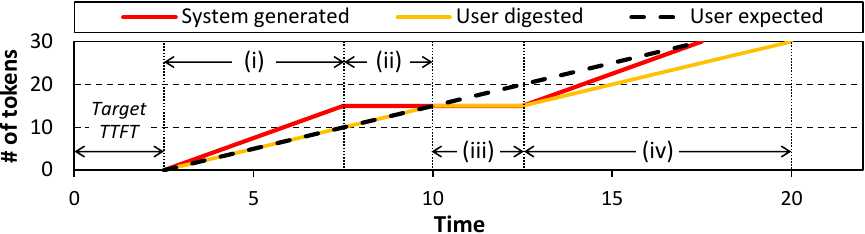}
    \caption{
    An example token serving scenario with QoE measurement, where token delivery begins at the target TTFT without additional delay. (i) The serving system initially generates tokens (red line) faster than the user’s expected reading pace (black dotted line), causing tokens to be buffered by the token pacer. The user consumes tokens at their own pace, represented by the user digested line (yellow). (ii) When the serving system is temporarily paused, the user continues consuming the buffered tokens. (iii) Once all the tokens kept in the buffer are depleted, the user experiences starvation until token generation resumes. (iv) Once token generation resumes, the user continues consuming tokens at a steady pace, although the user's token digestion timeline remains behind the expected schedule due to the earlier delay experienced when the serving system paused. QoE is measured by computing the ratio between the area under the user digested token curve (yellow) and the area under the user expected curve (black dotted line). 
    }\label{fig:background_qoe}
\end{figure}

An unfortunate side effect of time-sharing schedulers is that they can cause irregular token-generation rates. In \fig{fig:background_scheduling}(c), for instance, Request B is preempted for 3 time units, increasing the latency between its fourth and fifth tokens and potentially degrading user experience. Prior work~\cite{andes} addresses this issue using a \emph{token pacer}. The token pacer temporarily buffers tokens when they are generated in bursts and tries to release them at a steady rate whenever the request is idle due to preemption, thereby smoothing the output rate perceived by the user. To quantitatively measure how well the serving system's token generation rate aligns with the user’s token consumption rate, \cite{andes} introduced the \emph{Quality-of-Experience (QoE)} metric, which more accurately reflects SLO attainment than the conventional average TPOT measure. QoE combines two factors: (1) whether each request meets the target TTFT, and (2) how closely the effective token generation speed delivered to the user matches the target TPOT, accounting for the effects of token pacing. \fig{fig:background_qoe} illustrates how QoE is measured, which reflects how closely the token generation rate aligns with the user’s expected responsiveness. QoE is normalized to the interval [0,1] and a score of 1 indicates perfect alignment, while lower values suggest delays or starvation in token delivery (e.g., the example in \fig{fig:background_qoe} exhibits a QoE value less than 1 due to the lagging of token digestion timeline vs. the user expected timeline).

\subsection{{Rethinking TTFT and TPOT in Reasoning-Based LLMs}}
\label{sect:reasoning_models}

Reasoning-based LLMs generate more accurate responses and solve complex problems by incorporating techniques like Chain-of-Thought (CoT)\cite{cot_prompt, cot_tree, self_cot, least_cot}, in which models explicitly work through intermediate reasoning steps before arriving at final answers. This approach mimics human problem-solving by breaking down complex tasks into logical sequences. Over time, several reasoning algorithms have emerged, ranging from structured approaches such as tree-structured reasoning\cite{cot_tree} to explicit prompting methods~\cite{cot_prompt}. More recently, the field has converged toward methods in which the reasoning process is fully integrated into a single decoding pipeline, exemplified by models such as DeepSeek-R1~\cite{deepseekr1} and OpenAI o3~\cite{o4-mini}. These integrated reasoning-based LLMs automatically perform internal deliberation steps (i.e., the reasoning phase) within the same neural network architecture and inference process. They employ advanced training methods that enable the internalization of reasoning capabilities directly within the model architecture~\cite{deepseekr1, o4-mini}.

These fundamental differences from conventional LLMs affect how serving metrics should be interpreted. In conventional LLMs, TTFT and TPOT map cleanly to two distinct execution stages: TTFT captures the latency of the prefill stage, while TPOT measures how quickly output tokens are generated during the subsequent decoding stage. As illustrated in \fig{fig:ttft_and_tpot}(a), conventional LLMs generate the first user-visible token ($t_1$) immediately after the prefill stage, so TTFT spans only the grey-colored block. In contrast, reasoning-based LLMs internally divide the decoding stage into two sub-stages: (1) a \emph{reasoning phase}, which generates latent reasoning tokens not visible to the user, and (2) an \emph{answering phase}, producing the user-visible answering tokens. As shown in \fig{fig:ttft_and_tpot}(b), a reasoning-based LLM first generates reasoning tokens ($r_1$, $r_2$, and $r_3$) before the answering phase begins at $t_1$. As a result, the latency to generate these reasoning tokens contributes to TTFT, even though they occur during decoding. {That is, TTFT of reasoning-based LLMs consists of sum of the prefill-stage latency, the latency to execute the reasoning phase, and also the latency from the end of the reasoning phase to the generation of the first answering token (\fig{fig:ttft_and_tpot}(b)).} Consequently, the decoding stage now contributes to both TTFT {(during the reasoning phase)} and TPOT (during the answering phase), breaking the conventional one-to-one mapping of TTFT to prefill and TPOT to decode. For the remainder of this paper, we define the reasoning phase as encompassing both the prefill stage and the generation of reasoning tokens within the decoding stage, as both contribute to the latency before the first answering token becomes visible to the user.

\section{Characterization and Motivation}
\label{sect:characterization}

\subsection{Methodology}

This section characterizes how different scheduling policies impact reasoning-based LLM inference. We present key insights that inform efficient scheduling decisions for reasoning-based LLMs, with a particular focus on optimizing user experience. All experiments in this section were conducted on a server node hosted by Intel Xeon Platinum 8558 CPU (with {256 GB DDR5 DIMM}) communicating with an NVIDIA H100 GPU (with 96 GB HBM) communicating over PCIe 5.0. For LLM serving, we employ {vLLM v0.6.1~\cite{vllm}.

{\bf Workload.} We conducted experiments using DeepSeek-R1-Distill-Qwen-32B~\cite{deepseekr1, qwen32b}, an open-source LLM designed for reasoning tasks. To clearly characterize and motivate how different scheduling decisions affect the SLO metrics of the reasoning-based LLMs, 
we construct synthetic workloads as follows. Two separate experiments are conducted independently, each focusing on either the reasoning phase (\fig{fig:latency_characterization}) or the answering phase (\fig{fig:characterization}). In each experiment, our LLM serving system processes a total of 300 requests, with request arrivals following a Poisson distribution. The first experiment targets the reasoning phase (\fig{fig:latency_characterization}) where all requests have a fixed input prompt length of 128 tokens, while the reasoning token length is randomly selected between 128 and 2048 tokens. This setup allows us to isolate how different scheduling policies impact the performance of the reasoning phase under varying reasoning token lengths. In the second experiment characterizing the answering phase (\fig{fig:characterization}), we assume that all requests have already completed the execution of both prefill and the reasoning phase (the combined length of prefill and reasoning tokens is fixed at 128 tokens), and that their corresponding KV caches are already generated during previous execution. 
Under this setting, each request generates answering tokens with lengths ranging from 128 to 2048, which is chosen randomly, allowing us to analyze how different scheduling policies affect SLO attainment during the answering phase.

{\bf Scheduling policy.} To demonstrate how GPU memory constraints influence scheduling efficacy, we compare two scenarios: (1) an \emph{oracle} configuration with sufficient memory to hold the KV caches of all 300 requests simultaneously, allowing the scheduler to never have to block or preempt requests due to GPU memory limits, and (2) a memory‑constrained configuration. In our NVIDIA H100 (96 GB) GPU based system, the full‑capacity setting can accommodate all KV caches concurrently, so we treat it as the oracle configuration with no performance degradation attributable to memory limits. To emulate a realistic serving environment with GPU memory constraints, we cap the GPU memory available for KV cache allocation to 50\% of the oracle capacity and schedule requests based on FCFS or RR, which will block incoming requests or preempt ongoing ones once memory is exhausted (\fig{fig:background_scheduling}).

\subsection{Characterization}

\begin{figure}
    \centering
    \includegraphics[width=0.95\columnwidth]{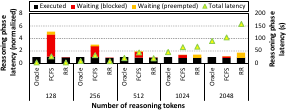}
    \caption{
    {(Left axis) Breakdown of average reasoning phase latency across oracle, FCFS, and RR scheduling policies for varying numbers of reasoning tokens (x-axis), normalized to the oracle latency at each reasoning token count. Executed (black) denotes time the request actively ran on the GPU without being blocked (red) or preempted (yellow).}\label{E-Q2} (Right axis) Corresponding average absolute latency in seconds, {indicated by green triangles.}
    }\label{fig:latency_characterization}
    \vspace{-2.0em}
\end{figure}

\begin{figure}[t]
    \centering
    \includegraphics[width=0.92\columnwidth]{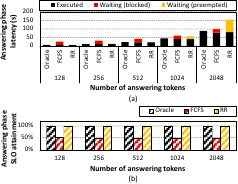}
    \caption{
        (a) Answering phase latency breakdown across varying numbers of answering tokens and (b) the corresponding SLO attainment rate.
        The SLO target for the answering phase is determined by two factors: (1) the latency between the end of the reasoning phase and the generation of the first answering token (henceforth referred to as \emph{Time-To-First-Answering-Token} (TTFAT)), and (2) whether the answering token generation rate meets the TPOT target.
        As such, per our definition of QoE in \fig{fig:background_qoe}, the answering phase's QoE is determined by the target values of TTFAT and TPOT.
        Following prior work~\cite{distserve}, {we set the target TTFAT and TPOT values to 0.25 seconds and 100 milliseconds, respectively.} A request violates the SLO if its QoE score falls below 0.95.
    }
    \label{fig:characterization}
    \vspace{-1.2em}
\end{figure}
{\bf Reasoning phase.}
\fig{fig:latency_characterization} illustrates how blocking and preemption aggravate reasoning phase latency relative to the oracle scenario (\fig{fig:background_scheduling}(a)). Unlike conventional LLMs where TTFT is determined solely by the prefill stage, reasoning-based LLMs produce the first (user-visible) answering token only after \emph{all} reasoning tokens are generated. Consequently, TTFT spans not only the prefill stage but also the entire decoding of reasoning tokens and can far exceed the sub‑second TTFT SLO typical of conventional LLMs, making timely completion of the reasoning phase critical for user satisfaction. \fig{fig:latency_characterization} therefore focuses on how scheduling policies shape reasoning phase latency under memory sufficiency (oracle) vs. memory pressure (FCFS, RR). As shown, both blocking (FCFS) and preemption (RR) add directly to reasoning phase latency, especially when limited GPU memory prevents uninterrupted execution. Under FCFS, blocking inflates latency because a request must wait behind long‑running ones whose KV caches remain resident until completion. This effect is most pronounced for short reasoning requests (e.g., 128 reasoning tokens), where wait time dominates and latency increases by up to $5.14\times$ over the oracle. In contrast, RR avoids outright blocking by interleaving requests, but frequent preemptions fragment execution. For long reasoning requests (e.g., 2048 reasoning tokens), RR increases latency by up to $1.75\times$ vs. uninterrupted oracle execution. Because RR enforces a fixed token quantum, long requests cannot retain continuous residency of their KV caches in GPU memory, accumulating repeated interruption and waiting intervals. Similar degradation is expected for other time‑sharing policies that necessarily preempt long requests to preserve fairness.

In summary, both blocking (FCFS) and preemption (RR and other time‑sharing schedulers) during the reasoning phase substantially delay generation of the first answering token under GPU memory constraints, severely degrading TTFT and motivating memory‑ and phase‑aware scheduling strategies.

{\bf Answering phase.}
\fig{fig:characterization} shows how scheduling policies affect user experience during the answering phase. \fig{fig:characterization}(a) shows a breakdown of the latency of answering phase across varying numbers of answering tokens and \fig{fig:characterization}(b) shows the corresponding SLO attainment rate. In reasoning-based LLMs, TTFT is measured as the latency until the ``first'' answering token is generated. Unlike the reasoning phase, where any additional delay directly increases TTFT, the answering phase can still meet its SLO if (1) the latency from the end of reasoning to the first answering token does not exceed the \emph{Time-To-First-Answering-Token} (TTFAT) target, which is set to be near-instantaneous to ensure an immediate transition from the reasoning phase to the answering phase, and (2) the steady-state answering token generation rate meets the TPOT target (\fig{fig:background_qoe})\footnote{
Our target TPOT was selected from a user-centric perspective rather than being tied to GPU performance or model specifications. This decision is grounded in the principle that user satisfaction is largely determined by whether the LLM produces output tokens at a rate comparable to typical human reading speeds. In other words, even if future GPUs provide 10$\times$ greater compute capability, we do not expect TPOT to decrease proportionally, as there is little benefit in ``over-optimizing'' it beyond the threshold of human perception. Likewise, even with future LLMs that demand 10$\times$ more computation, we expect TPOT to remain within the 100 ms range. Prior studies~\mbox{\cite{throttling, mlperf, andes, deepspeed_fastgen}} have similarly argued that a TPOT of around 100 ms (which is already faster than human reading speeds) is sufficient to provide a natural, conversational experience.
}.  Because SLO satisfaction is threshold-based rather than minimizing absolute latency (i.e., latencies only need to be \emph{fast enough}), appropriate scheduling can preserve user experience even when execution is fragmented by blocking or preemption. 

Under FCFS, blocking forces requests to wait behind long-running ones (\fig{fig:characterization}(b)), substantially increasing {TTFAT} and lowering SLO attainment across all answering token lengths (\fig{fig:characterization}(b)). In contrast, RR eliminates head-of-line blocking by interleaving execution, allowing all requests to start promptly to keep {TTFAT} low, while also sustaining adequate token generation rate. Consequently, SLO attainment remains high for most configurations with RR. Notably, at 2048 answering tokens, RR incurs higher total answering phase latency than FCFS due to preemptions (\fig{fig:characterization}(a)), yet achieves the same SLO attainment as the oracle because both its {TTFAT} and per-token generation rate remain within thresholds. This shows that during the answering phase, time-sharing policies like RR can tolerate preemption overhead while still preserving user experience, unlike their detrimental impact during the reasoning phase, as highlighted in \fig{fig:latency_characterization}.

In summary, answering phase SLOs are robust to moderate RR-induced preemption overhead as long as threshold metrics are met, making time-sharing schedulers like RR preferable to blocking policies like FCFS for this phase.

\subsection{Motivation}

Our analysis reveals that reasoning and answering respond very differently to blocking and preemption. In the reasoning phase, blocking or preemption directly stretches reasoning latency, delaying the generation of the first answering token and degrading TTFT-driven user experience. In contrast, limited preemption does not necessarily harm user experience in the answering phase. As long as the first answering token appears promptly after reasoning phase's completion and the steady-state answering token generation rate meets the required throughput threshold, SLOs remain satisfied even if execution is fragmented. Time‑sharing policies (e.g., RR) therefore mitigate head‑of‑line blocking and preserve answering-phase {QoE, despite incurring preemption overhead}.

In summary, two key insights emerge. (1) The reasoning and answering phases exhibit fundamentally different sensitivities: reasoning latency is highly interruption‑sensitive, while answering QoE is threshold‑ (not absolute latency‑) sensitive, and (2) time‑sharing is especially effective in the answering phase because it avoids blocking without violating SLO thresholds.

Existing scheduling policies targeting conventional (single‑phase) LLM inference ignore the distinct latency vs. threshold sensitivities of reasoning-based workloads. This gap motivates phase‑aware scheduling that adapts its strategy, minimizing interruptions during reasoning while employing fair time‑sharing during answering to jointly optimize reasoning latency and answering phase user experience.

\section{\proposed: A Phase-Aware Scheduling Algorithm for Serving Reasoning-based LLMs}
\label{sect:proposed}

\proposed is a scheduling system that improves the user experience for reasoning-based LLMs on multi-instance LLM serving systems. An instance is the software abstraction in LLM serving frameworks that serves as the unit of execution, managing a replica of the model weights and coordinating access to shared GPU resources. \proposed schedules inference requests differently in the reasoning and answering phases by exploiting their asymmetric sensitivity to preemption. Specifically, it minimizes interruptions during the reasoning phase to reduce TTFT and permits controlled preemption during the answering phase without compromising SLO goals for TPOT.

\begin{figure}
    \centering
    \includegraphics[width=0.9\columnwidth]{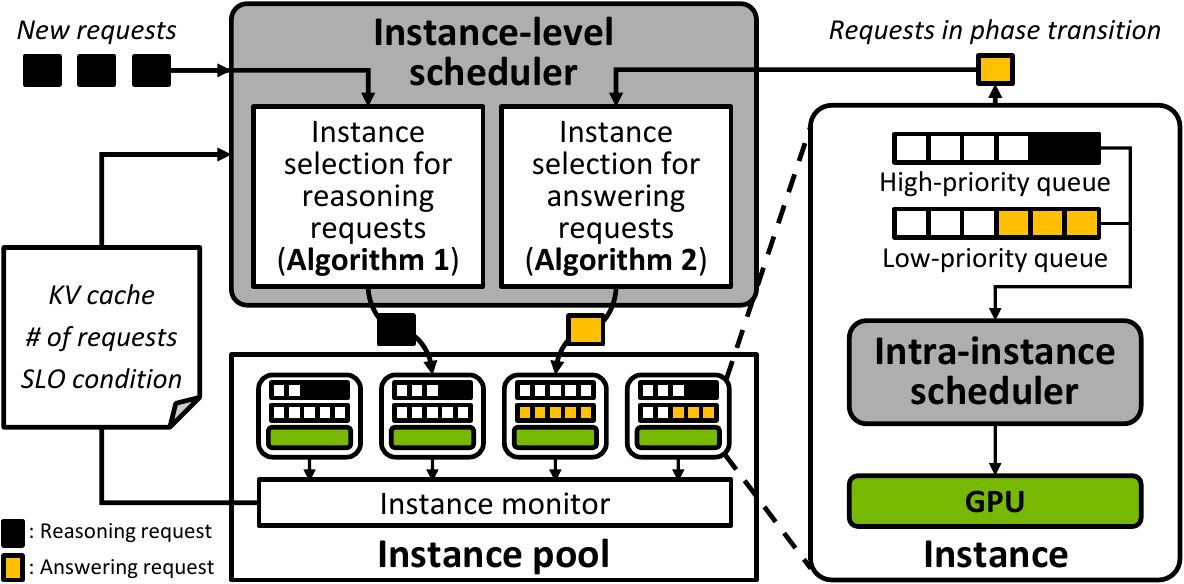}
    \caption{
Overview of \proposed. The system comprises an instance-level scheduler and a pool of serving instances. All incoming requests first enter the reasoning phase and are routed to the appropriate instances using \algo{alg:reasoning-selection}. After a request transitions to the answering phase, the instance-level scheduler selects a (possibly different) instance using \algo{alg:answering-selection}. Within each serving instance, an intra-instance scheduler manages routed requests through high- and low-priority queues and orchestrates their execution. An instance monitor continuously collects runtime metrics to guide these placement decisions.    }\label{fig:overview}
    \vspace{-1.2em}
\end{figure}

\subsection{High-level Overview}
\label{sect:system_overivew}

We take a top-down approach to explain \proposed's two-level scheduling architecture (\fig{fig:overview}).  The \emph{instance-level scheduler} selects which GPU instance should service each incoming request. The \emph{intra-instance scheduler} then orchestrates request batching and kernel execution within the instance.

The instance-level scheduler employs two placement algorithms, one for each phase of a request: reasoning and answering. When a request first arrives, it enters the reasoning phase\footnote{For clarity, we collectively refer to the prefill stage and the subsequent reasoning phase of decoding as the reasoning phase.}. During this phase, the request generates intermediate reasoning tokens from the input prompt until reasoning completes. For requests in this phase, the scheduler selects the instance that minimizes reasoning latency while preserving the performance of other requests already executing on each GPU instance. After the system’s instance monitor detects that the most recent token generated by an existing request is the special token signaling the end of the reasoning phase (e.g., the \texttt{<\textbackslash think>} token {in DeepSeek-R1}), the scheduler re-examines whether migrating that request to a different instance would be beneficial. Such migration aims to minimize any delay after reasoning completes, satisfy answering-phase SLOs, and reduce interference with co-located requests.

Based on the instance-level scheduler’s decisions, a single GPU instance may host requests from both the reasoning and answering phases. To manage these mixed phase requests effectively, \proposed deploys a local intra-instance scheduler that coordinates execution across phases. It prioritizes the scheduling of reasoning requests by placing them in a high-priority scheduling queue, since their latency directly affects TTFT and is highly sensitive to blocking and preemption. Answering requests are inserted to a low-priority queue, and the scheduler time-shares the remaining GPU resources among them and seeks to preserve a consistent user experience. {Requests within the same (high-/low-priority) queue are scheduled based on a round-robin policy.}

\subsection{Instance-level Scheduler}
\label{sect:load_balancing}

The instance-level scheduler selects the GPU instance for each request based on its current phase and SLO requirements. Consequently, the scheduler uses two distinct instance-selection algorithms, one per phase, each conditioned on the current status of an instance’s request queues. As noted earlier, reasoning requests are routed to the high-priority queue within a GPU instance, whereas answering requests are routed to the low-priority queue. Accordingly, \proposed employs two algorithms: one for assigning new reasoning requests (\algo{alg:reasoning-selection}) and another for determining whether requests that have entered the answering phase should be migrated to a different instance (\algo{alg:answering-selection}). These algorithms consider each instance’s queue occupancy and the characteristics of individual requests, aiming to balance memory usage while respecting the phases’ differing latency sensitivities.

\begin{algorithm}[t!]
\footnotesize
\caption{Instance Selection for Reasoning Requests}
\label{alg:reasoning-selection}
\begin{algorithmic}[1]
\State \textbf{Input:} Set of instances $I$, where for each instance $i \in I$, we have:
\Statex \quad $t_i$: whether all answering requests meet their SLOs (TRUE/FALSE)
\Statex \quad $m_i$: total memory occupied by KV cache (GPU + CPU)
\State \textbf{Output:} Selected instance {to schedule} the reasoning phase request
\State $E \gets \{\, i \in I \mid t_i = \text{TRUE} \,\}$
\If{$E = \emptyset$}
    \State $E \gets I$
\EndIf
\State $selected \gets \operatorname{argmin}_{i \in E}(m_i)$
\State \Return $selected$
\end{algorithmic}
\end{algorithm}

{\bf Instance selection for reasoning requests.}
As described in \algo{alg:reasoning-selection}, the instance-level scheduler first inspects each GPU instance’s token pacer to determine whether answering requests are being served at the expected user-perceived rate. If the token pacer reports insufficient remaining tokens (indicating token generation is slower than the user’s expected pace), the instance is deemed to be violating its answering-phase SLO and is excluded from the candidate set (i.e., $t_i = \text{FALSE}$ in line 3).
Recall that answering requests are assigned a  lower priority than the reasoning requests and thus must rely on whatever GPU memory remains once the high-priority queue has allocated KV cache space for reasoning requests. Therefore, an instance already missing its answering-phase SLO implies that available GPU memory is insufficient to support the current answering requests. Consequently, scheduling additional high-priority reasoning requests to such instance would only intensify memory pressure and further delay answering requests. Among the remaining SLO-compliant candidate instances, the scheduler selects the instance with the smallest KV cache footprint $m_i$ (line 7). A smaller KV cache footprint generally indicates fewer active requests, making the instance more amenable to accommodating additional new requests without {interfering with requests already under execution}. Moreover, smaller KV cache correlates with shorter attention-layer execution time, increasing the likelihood of meeting TTFT for new reasoning requests, further justifying this design decision.  When no instance meets the SLO condition (i.e., $E=\emptyset$ in \algo{alg:reasoning-selection}), the scheduler still chooses the instance with the minimum $m_i$ to minimize additional performance degradation for in-flight requests (lines 4–5).

\begin{algorithm}[t!]
\footnotesize
\caption{{Instance Selection for Answering Requests}}
\label{alg:answering-selection}
\begin{algorithmic}[1]
\State \textbf{Input:} Set of instances $I$, where for each instance $i \in I$, we have:
\Statex \quad $t_i$: whether all answering requests meet their SLOs (TRUE/FALSE)
\Statex \quad $r_i$: number of reasoning requests {in high-priority queue}
\Statex \quad $a_i$: number of answering requests {that has not exhausted the first time quantum} 
\State \textbf{Output:} Selected instance {to schedule the} answering phase request
\State $E \gets \{\, i \in I \mid t_i = \text{TRUE} \,\}$
\If{$E = \emptyset$}
    \State $E \gets I$
\ForAll{$i \in E$}
    \State $r_i \gets r_i + a_i$
\EndFor
\EndIf
\State $selected \gets \operatorname{argmin}_{i \in E}(r_i)$
\State \Return $selected$
\end{algorithmic}
\end{algorithm}

{\bf Instance selection for answering requests.} \label{proposed_answering_selection} During the execution of a reasoning request, the instance monitor in \fig{fig:overview} continuously checks for the special token (e.g., \texttt{<\textbackslash think>}) that marks the transition to the answering phase. When this token appears (signaling that the model is ready to emit answer tokens), the instance-level scheduler migrates the request to a new GPU instance according to \algo{alg:answering-selection}, {unless our adaptive migration policy (discussed further at the later part of this subsection) overrides such migration decision}. As with {\algo{alg:reasoning-selection}}, any instance that is currently failing to meet answering-phase SLOs (i.e., $t_i=\text{FALSE}$) is removed from the candidate set (line 3). From the remaining candidates, the scheduler selects the instance with the fewest active reasoning-phase requests ($r_i$) to minimize the chance that the answer request is delayed by higher-priority reasoning requests (line 10). Because reasoning requests reside in the high-priority queue and allocate GPU memory first, an instance hosting many such requests often leaves too little GPU memory for a new answering request. Insufficient GPU memory forces answering requests either to spill their KV cache to CPU memory or to wait until resources are freed—both of which compromise answering-phase SLOs.
This motivates selecting the instance with the smallest $r_i$ for answering requests.

If no instance satisfies the SLO condition, the scheduler selects the one with the smallest $r_i + a_i$ (lines 4--9), where $a_i$ denotes {the number of answering requests in the low-priority queue that have not yet exhausted the first time quantum (i.e., they are more likely to be scheduled next under the RR policy)}. As described in Section~\ref{sect:background}, RR scheduling is fairness-oriented; the more ``fresh'' answering requests ($a_i$) an instance holds, the more the newly arriving answering request must compete for scheduling opportunities, an undesirable outcome when selecting the instance for new answering requests.  Our empirical evaluation confirms that \proposed's {such heuristic based instance-selection algorithm}, which considers both $r_i$ and $a_i$, achieves better load balancing and SLO attainment than using $r_i$ alone under these scenarios.

\begin{figure}
    \centering
    \includegraphics[width=0.92\columnwidth]{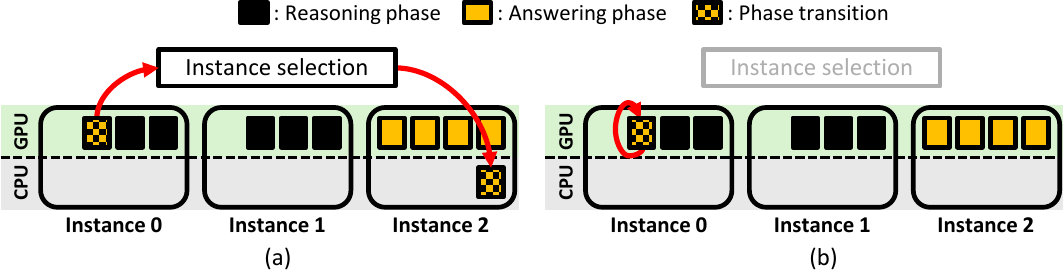}
    \vspace{-0.5em}
    \caption{
Adaptive migration example. (a) \algo{alg:answering-selection} initially recommends migrating the phase transitioning request to Instance 2. Because GPU memory in Instance 2 is fully utilized (indicated by four yellow blocks) while Instance 0 still has free memory (two empty slots), (b) the system overrides \algo{alg:answering-selection}'s migration decision and continues executing the phase transitioning request in Instance 0.
}
    \label{fig:load_adaptive}
    \vspace{-1.1em}
\end{figure}

{\bf Adaptive migration.} Although requests are generally migrated to a new instance when they transition to the answering phase, strictly following \algo{alg:answering-selection}'s decision can lead to suboptimal resource utilization in certain scenarios. Therefore, \proposed introduces an \emph{adaptive} migration mechanism that determines whether to migrate a request during phase transitions based on the current memory availability. This strategy {becomes particularly important in} scenarios where reasoning and answering requests are distributed asymmetrically across instances. \fig{fig:load_adaptive}(a) shows one such case when reasoning requests are distributed across many instances, while answering requests are concentrated on a certain instance. Since answering requests are directed to instances with the fewest active reasoning requests (\algo{alg:answering-selection}), they tend to accumulate on the same instance (\texttt{Instance 2}). In this case, when a reasoning request in \texttt{Instance 0} transitions to the answering phase, \algo{alg:answering-selection} likely selects \texttt{Instance 2} based on the number of reasoning requests, which lacks available GPU memory. Despite \texttt{Instance 0} having sufficient memory to handle the answering phase, migrating the request to \texttt{Instance 2} can cause the unnecessary execution stall for answering requests in \texttt{Instance 2}. To handle this case, at phase transition, if the current instance has sufficient GPU memory while the one selected for answering phase does not, \proposed preserves the request on the current instance rather than migrating it to the selected one (\fig{fig:load_adaptive}(b)). This avoids unnecessary CPU offloading and KV cache transfer overhead, and it better utilizes available GPU memory.

{\bf KV cache transfer overhead.} 
{Migrating a request at phase transition requires transferring its entire KV cache from the source to the destination instance.} Unlike prior work~\cite{splitwise, distserve, llumnix}, which can overlap KV cache transfers with computation, reasoning-based LLMs cannot predict phase transitions in advance, as transitions only become apparent when the model emits specific tokens (e.g., \texttt{<\textbackslash think>}) that indicate the end of the reasoning phase. In reasoning-based LLMs, this transfer latency can impact TTFT because the server must wait for both the completion of the reasoning phase and the subsequent KV cache migration \emph{before} generating the first answering token. 
Nonetheless, we find that the overhead of KV cache transfers is negligible in reasoning models, as TTFT is primarily dominated by reasoning phase latency. For example, Patel et al.\cite{splitwise} analyzed KV cache transfer latency in a disaggregated LLM serving system and reported a one‑time transfer delay of approximately 40$\mathrm{ms}$ for 2,048 tokens.  To put this in context, consider a per-token latency of 30$\mathrm{ms}$~\cite{splitwise}, which reflects an aggressive decoding speed. Because reasoning phases can involve hundreds or even thousands of tokens—equating to several tens of seconds of reasoning latency—the one‑time KV cache transfer constitutes only a small fraction of the total TTFT and is therefore negligible.

\subsection{Intra-instance Scheduler}
\label{sect:hierarchical_priority}

This subsection explains how the intra-instance scheduler manages and executes requests within each GPU instance. As shown in \fig{fig:overview}, each instance maintains a hierarchical priority queue that separates the reasoning and answering phases based on their distinct performance requirements:

\begin{itemize}
\item {\bf High-priority queue} stores requests in the reasoning phase. These requests are served first and receive preferential allocation privilege to GPU memory, minimizing the impact of blocking or preemption on TTFT.

\item {\bf Low-priority queue}: stores requests in the answering phase. Because these requests tolerate preemption better, they are executed using whatever GPU memory remains after high-priority allocations are satisfied.

\end{itemize}

{
Occasionally, a single reasoning request with an exceptionally long sequence and a very large KV cache can exhaust the memory needed by answering requests. In such cases, the intra-instance scheduler \emph{demotes} the reasoning request whose KV-cache size exceeds a user-defined threshold to the low-priority queue, effectively treating it as a low-priority answering request. This conditional demotion policy frees GPU memory for answering requests more quickly while minimally impacting reasoning-phase SLOs.}

For each priority queue, the scheduler applies a different dispatch strategy. For reasoning requests in the high-priority queue, \proposed adopts a RR policy. As discussed in \fig{fig:latency_characterization}, when the serving system is constrained by GPU memory, offloading even high-priority reasoning requests becomes inevitable, and both FCFS and RR experience latency degradation. However, the severity of this degradation differs: FCFS increases average latency due to head-of-line blocking across all request lengths, whereas RR provides much faster responsiveness for short reasoning requests at the cost of higher tail latency (\fig{fig:latency_characterization}). To this end, \proposed selects RR for the high-priority queue to guarantee lower TTFT for short reasoning requests under memory pressure.

Unlike reasoning requests, tokens generated during answering requests are streamed directly to users, making perceived responsiveness critical. As evaluated in \fig{fig:characterization}, RR exhibits high robustness to preemption‑induced latency overheads and achieves high SLO attainment rates. To mitigate user‑experience degradation under preemption or delayed execution, \proposed augments RR with a token pacer for answering requests in the low‑priority queue. The token pacer regulates the token emission rate to match user expectations, allowing even preempted answering requests to appear responsive. This mechanism enables \proposed to maintain a high answering‑phase SLO success rate under constrained memory, outperforming an FCFS baseline.

\section{Evaluation}
\label{sect:evaluation}

\begin{figure}
    \centering
\includegraphics[width=0.94\columnwidth]{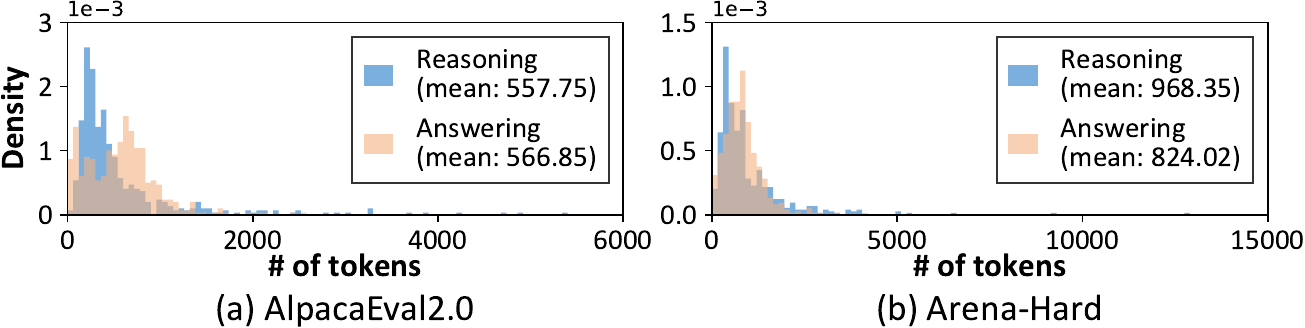}
    \caption{{Reasoning and answering token count distribution for (a) AlpacaEval2.0~\cite{alpaca} and (b) Arena-Hard~\cite{arena} datasets.}}
    \label{fig:token_distribution}
    \vspace{-1.2em}
\end{figure}

\subsection{Methodology}
\label{sect:eval_method}

{\bf Simulation framework.}  We developed a cluster‑level simulator to evaluate design points enabled by \proposed. Inspired by the methodology in{~\cite{splitwise},} our simulator models eight server nodes (instances) connected via a 100 Gbps fabric, each with an NVIDIA H100 96 GB GPU. For single serving instance simulation, we adopt a \emph{profile‑based} LLM serving methodology, which prior work has shown to strike a balance between simulation speed and accuracy~\cite{vidur, splitwise, vtrain, apex}. The single‑instance simulator uses vLLM profiling data and mirrors established techniques~\cite{vidur, splitwise, vtrain, apex} to capture GPU execution behavior. We validate it by comparing simulated and measured inference latency on a real system with an Intel Xeon Platinum 8558 CPU and H100 GPU (PCIe 5.0, vLLM 0.6.1, PyTorch 2.4.0, CUDA 12.1). Our simulator achieves a mean absolute percentage error (MAPE) of 1.62\% for end‑to‑end latency and captures key metrics with MAPE values of 12.6\% for mean TTFT and 6.49\% for TPOT.

{\bf Model and dataset.}
We evaluate \proposed using DeepSeek-R1-Distill-Qwen-32B~\cite{deepseekr1, qwen32b}, an open-source reasoning LLM. The 32B model was chosen to reflect LLM serving scenarios in which GPU memory constraints limit the allocation of KV caches across diverse requests, forcing the scheduler to block or preempt requests as necessary. Serving traces are built from AlpacaEval2.0~\cite{alpaca} and Arena-Hard~\cite{arena} prompts, queried via OpenAI’s o4-mini~\cite{o4-mini} API, which provides reasoning/answering token counts (\fig{fig:token_distribution}). These two benchmarks, also used in DeepSeek-R1’s evaluation~\cite{deepseekr1}, capture chat applications requiring detailed responses. We additionally analyze scenarios with long reasoning but short answers (e.g., problem-solving) in~\sect{eval:ablation}.

{\bf Scheduling policy.}
We compare \proposed against two baseline scheduling algorithms as follows:
\begin{itemize}
\item \textbf{FCFS}: The default policy in vLLM. Requests are served strictly in arrival order without distinguishing reasoning and answering phases. When KV cache usage exceeds GPU memory, the KV cache spill to CPU memory, and new admissions pause until space frees up.
\item \textbf{Round‑robin (RR)}: A preemptive scheduler that allocates fixed token quanta to active requests in circular order, regardless of which phase the request is executing; requests that finish their quantum are deprioritized.
\end{itemize}

For both baselines, reasoning requests are placed on the instance with the smallest KV footprint, and no migration occurs at phase transitions. Token quantum is set to 500 for RR and for each queue in \proposed. In \proposed, reasoning requests exceeding 5000 tokens are demoted to the low‑priority queue.

{\bf Metric.} We evaluate schedulers using two metrics: \textit{TTFT} and \textit{SLO violation}. TTFT measures latency from request submission to the first answering token. SLO violations are assessed via QoE, calculated from TPOT starting at the first answering token. Because reasoning-based LLMs have highly variable reasoning lengths, the original QoE metric (which includes a fixed TTFT target) is impractical; we instead compute QoE solely from TPOT and evaluate TTFT separately. An SLO violation is counted when QoE falls below 0.95.

\subsection{{User Experience (TTFT, SLO Violation) and Throughput}}
\label{sect:results}

\begin{figure}[tbp]
    \centering
    \includegraphics[width=0.98\columnwidth]{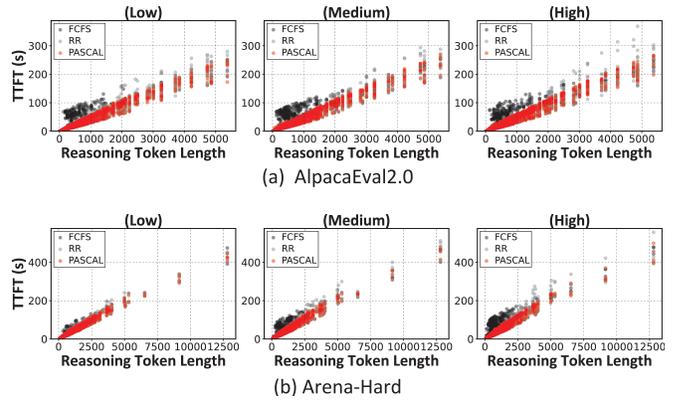}
    \caption{Absolute TTFT for all scheduled requests as a function of reasoning‑token length. Experiments were conducted under three request‑arrival rates {(low, medium, and high)}. The “high” rate stresses the LLM serving system more severely; exceeding GPU compute and memory capacity increases the likelihood of preemption and blocking. Request arrivals follow a Poisson distribution. 
    }
\label{fig:main_result_alpaca}
\end{figure}

\begin{figure}[tbp]
    \centering
    \includegraphics[width=0.88\columnwidth]{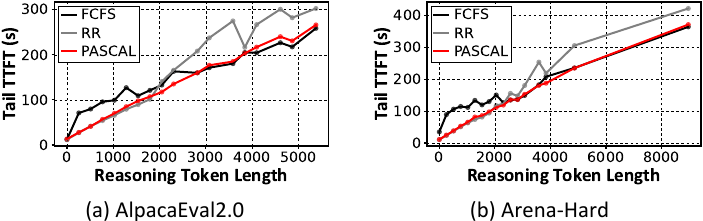}
    \caption{
    {Tail TTFT latency as a function of reasoning token length under {high} request-arrival rates. Requests are grouped into 256‑token bins based on reasoning token length (e.g., {[0–255] bin, [256–511] bin}, $\ldots$). These workloads exhibit a highly skewed distribution where more than {70\%} of requests generate fewer than 1,000 reasoning tokens. This skew causes substantial variation in the number of requests that fall under a given bin. Consequently, we use different tail latency metrics based on the sample size of each bin: maximum TTFT is used for bins with fewer than 10 samples, 90th-percentile (P90) for bins with fewer than 20 samples, P95 for bins with fewer than 100 samples, and P99 otherwise. Bins with fewer than five samples are omitted as we believe they are statistically less meaningful.
    }
}
    \label{fig:main_result_arena}
    \vspace{-0.7em}
\end{figure}

{\bf TTFT.} {
\fig{fig:main_result_alpaca} shows the raw TTFT values of all requests as a function of reasoning token length and \fig{fig:main_result_arena} summarizes the resulting \emph{tail} TTFT. Below we focus on tail TTFT because most requests fit within GPU memory and scheduling effects are most pronounced in the tail, where memory pressure triggers cache spills and} preemption.

The baseline FCFS suffers from head‑of‑line blocking because once a request starts reasoning, it holds GPU memory until its answering phase completes. As a result, even short‑reasoning requests experience severe tail TTFT degradation. {RR can mitigate head‑of‑line blocking, which improves TTFT for shorter requests. In RR, answering phases always follow reasoning phases within each request’s lifecycle, but they receive progressively lower priority as the request consumes more time quanta. This behavior implicitly creates a per‑request hierarchy (i.e., scheduling priority) that favors reasoning, similar to \proposed.} However, RR cannot enforce a global ``reasoning‑first'' scheduling priority as it does not distinguish the reasoning phase vs. answering phase.  For example, a short request that quickly enters its answering phase can retain higher priority than a long reasoning request that has already consumed many quanta, rendering RR's fairness‑oriented policy to deprioritize the longer request. \proposed maintains a consistent phase-aware scheduling policy across all requests: reasoning phases always preempt answering phases, and requests migrate to the instance with the least reasoning requests at phase transitions. This avoids head‑of‑line blocking and keeps tail TTFT low across all datasets and traffic loads. Compared to FCFS, \proposed cuts tail TTFT by up to 61\% on AlpacaEval2.0 and 72\% on Arena‑Hard, with absolute tail TTFT reductions of 43.22 sec and 64.21 sec, respectively. Compared to RR, \proposed reduces tail TTFT by up to 89.91 sec and 72.73 sec, a 33\% and 29\% reduction thereby significantly improving user experience.

It is worth noting that there exists a small number of cases where \proposed causes minor TTFT degradation, but the absolute increases are small. For example, in AlpacaEval2.0 the worst case is a 13.30 sec increase vs. FCFS (6.12\% higher TTFT) and an 8.25 sec increase vs. RR (9.23\% higher TTFT). This slight degradation is due to increased preemption among answering requests competing for limited GPU memory. Outside these rare outliers, \proposed consistently maintains lowest TTFT compared to all baseline schedulers.

\begin{figure}
    \centering
    \includegraphics[width=0.94\columnwidth]{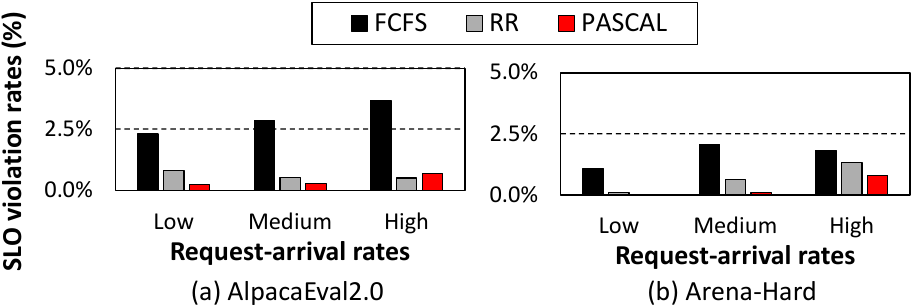}
    \vspace{-0.5em}
    \caption{
    {SLO violation rates across different request-arrival rates.}   
    }\label{fig:tpot_violation}
\end{figure}

{\bf SLO violations.} As shown in \fig{fig:tpot_violation}, \proposed consistently achieves {lower or comparable} answering‑phase SLO violation than both baselines across all request‑arrival rates. This advantage stems from \proposed's SLO‑aware instance‑level scheduling, which routes answering requests to instances with minimal reasoning load, while also prioritizing reasoning phase through its hierarchical queue.
At request‑arrival rates higher than those in \fig{fig:tpot_violation}, all schedulers inevitably see higher SLO violations, as no serving system can sustain performance once demand exceeds its resource capacity. However, \proposed confines most degradation to SLO violations while keeping TTFT low, whereas the baseline schedulers degrade in both TTFT and SLO metrics. This resilience arises from \proposed's strict reasoning‑phase prioritization via its hierarchical scheduler.

\begin{figure}
    \centering
    \includegraphics[width=0.94\columnwidth]{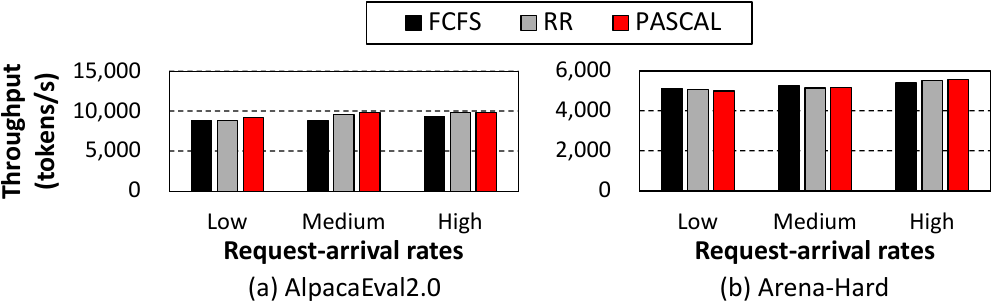}
    \caption{
    {Serving throughput across different request-arrival rates.}
    }\label{fig:throughput}
\end{figure}

{\bf Serving throughput.} \fig{fig:throughput} shows the overall throughput of the LLM serving system across various request-arrival rates and datasets. Throughput is measured as the total time required to generate all output tokens, including both reasoning and answering tokens. As shown, \proposed achieves throughput comparable to both baseline schedulers, differing by no more than 3\%. This result demonstrates that software‑level scheduling decisions that account for the distinct reasoning and answering phases of reasoning-based LLMs can significantly improve user experience (in terms of TTFT and TPOT) without compromising overall throughput.

\subsection{KV Cache Transfer Overhead}
\label{eval:kv_transfer}

In our multi‑instance server environment, multiple instances may simultaneously migrate requests transitioning into the answering phase (along with their associated KV caches) to the same target instance, causing bandwidth contention and potentially increasing TTFT and overall latency. In our evaluation, we observed P99 KV cache transfer latencies of 0.14 sec for AlpacaEval2.0 and 0.25 sec for Arena‑Hard under high request‑arrival rates. Since TTFT in these scenarios ranges from a few seconds to several hundred seconds, the impact of KV cache transfer overhead on end‑to‑end latency is negligible. These results demonstrate that, although KV cache migration introduces some overhead, the performance gains from intelligently migrating phase‑transitioning requests across instances far outweigh its costs.

\begin{figure}[t]
    \centering
    \includegraphics[width=0.98\columnwidth]{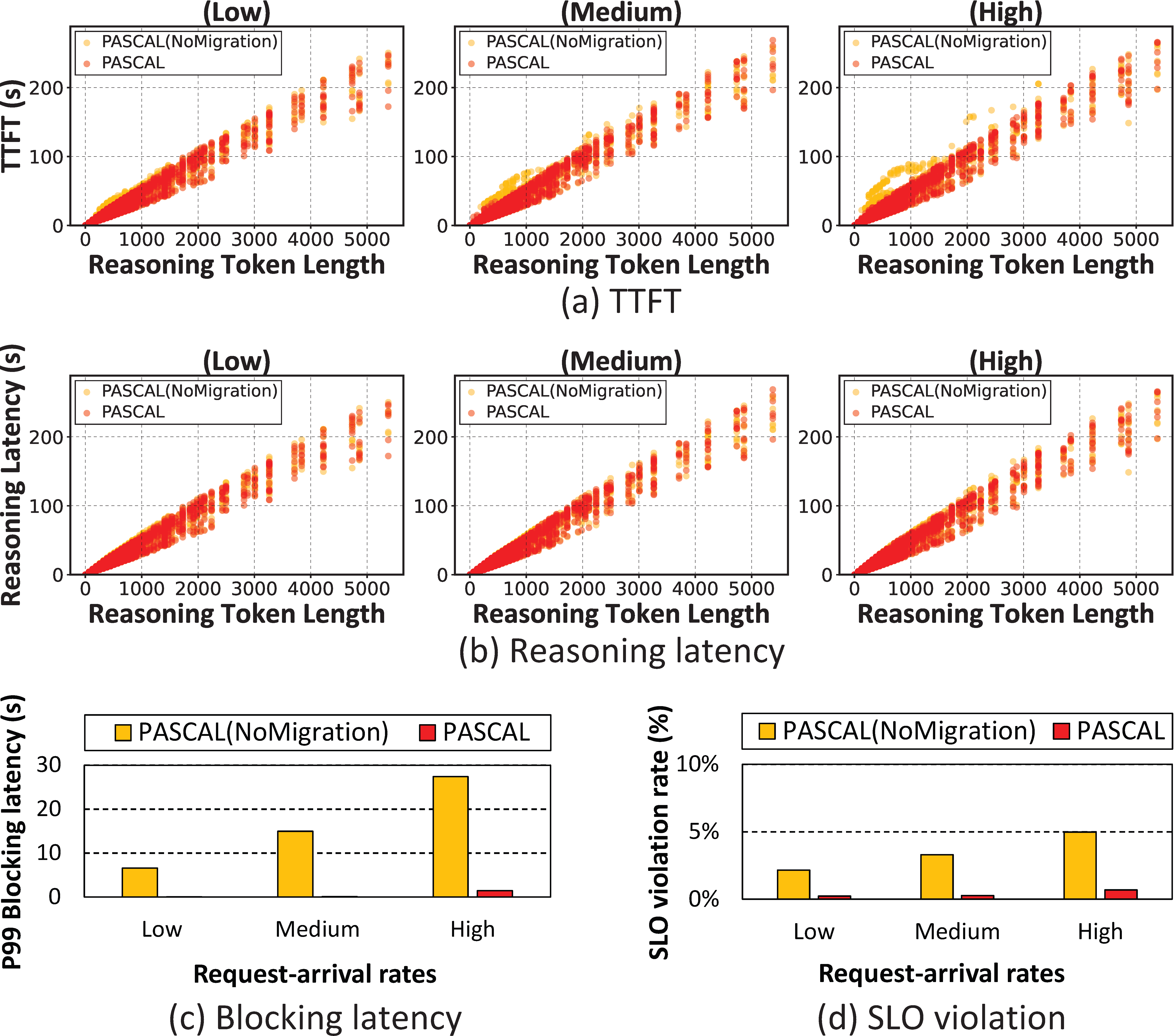}
    \caption{
    {
Evaluation of (a) TTFT, (b) reasoning latency, (c) P99 blocking latency, and (d) SLO violation rate on the AlpacaEval2.0 dataset when \proposed's inter‑instance request migration feature is disabled (\nomigration).
    }
    }\label{fig:hierarchical}
\end{figure}

\subsection{Discussions}
\label{eval:ablation}

{\bf Importance of migrating requests.}
To demonstrate the importance of migrating a request at the reasoning‑to‑answering boundary, we construct a variant of \proposed that \emph{disables} migration. This version (aka \nomigration) retains the hierarchical queue design but pins every request to the GPU instance chosen by \algo{alg:reasoning-selection}; once the reasoning phase ends, no inter‑instance migration occurs. Under \nomigration, the scheduler still prioritizes reasoning requests, but requests can still stall at the phase transition point. If GPU memory is fully utilized by high‑priority reasoning requests, an answering request that is routed to the same instance’s low‑priority queue cannot get scheduled until the GPU memory is freed. Because \proposed's original placement policy only considers the KV cache footprint during reasoning, it neglects the memory required for answering, thereby elevating the risk of SLO violations.

\fig{fig:hierarchical} illustrates these effects. \nomigration worsens tail TTFT under high request‑arrival rates across a broad range of reasoning‑token lengths (\fig{fig:hierarchical}(a)), even though reasoning phase latency remains almost unchanged (\fig{fig:hierarchical}(b)). To understand the cause, \fig{fig:hierarchical}(c) reports the latency between the moment a request transitions from reasoning to answering and the moment it is first scheduled. The P99 of this latency (referred to as \emph{blocking latency}) reaches up to 27.39 sec in \nomigration, whereas \proposed keeps it near zero. {Overall, \nomigration suffers from markedly higher answering phase SLO violation rates than \proposed (\fig{fig:hierarchical}(d)), highlighting the value of \proposed's SLO‑aware request migration at phase boundaries.}

\begin{figure}[t]
    \centering
    \includegraphics[width=0.98\columnwidth]{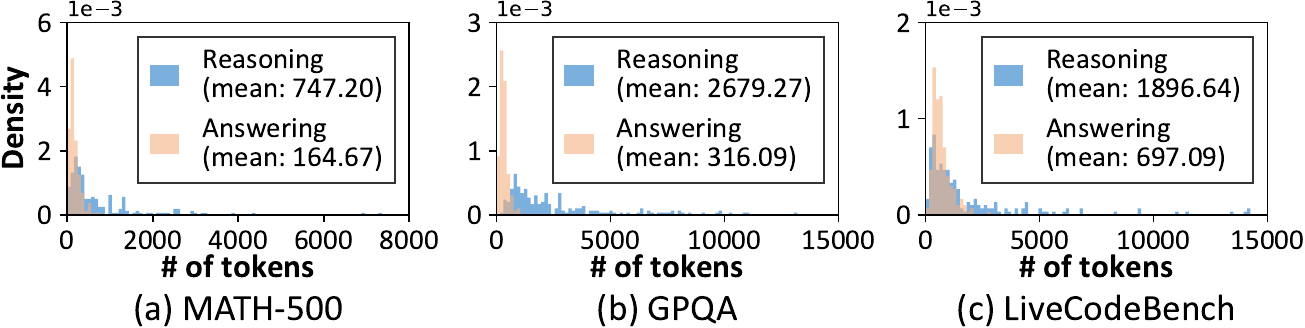}
    \caption{
    {
Reasoning and answering token count distributions for (a) MATH‑500~\cite{math-500}, (b) GPQA~\cite{gpqa}, and (c) LiveCodeBench~\cite{livecodebench}. Distributions are derived by querying each benchmark’s prompts into the o4‑mini model, following the methodology in \sect{sect:eval_method}.
}
}
    \label{fig:token_distribution_discussion}
\end{figure}

{
\begin{figure}
    \centering
    \includegraphics[width=0.98\columnwidth]{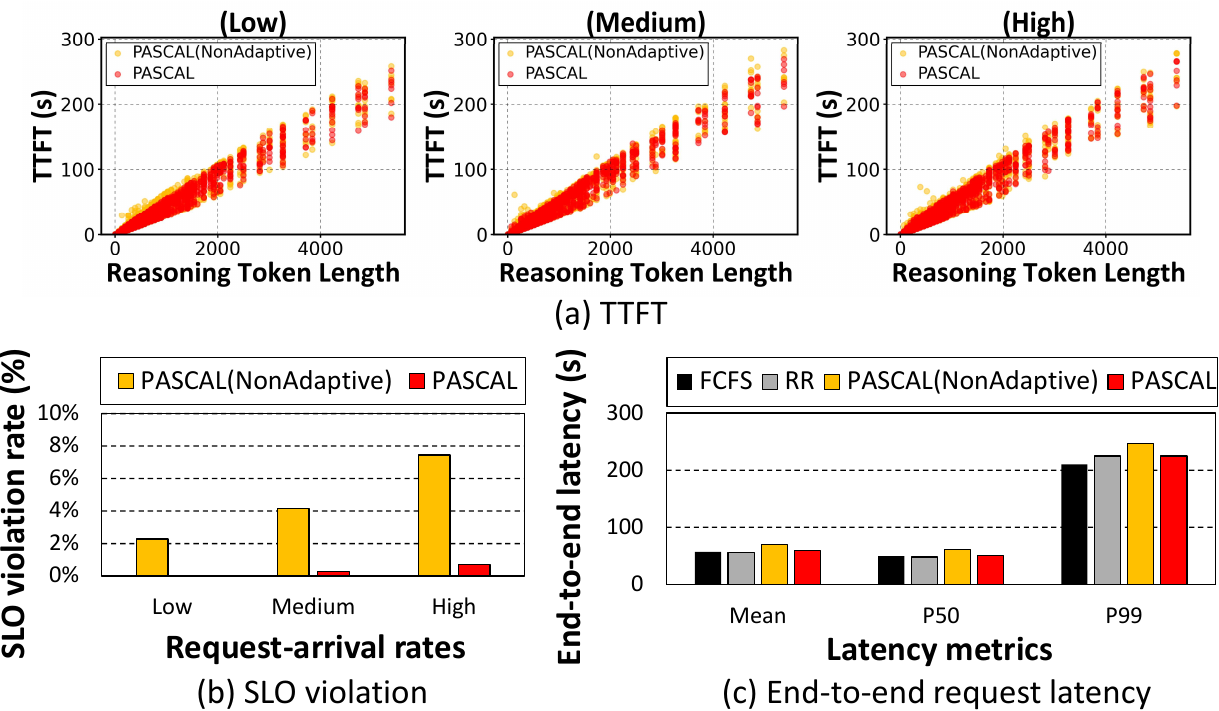}
    \caption{Comparison of \mbox{\nonadaptive} and \mbox{\proposed} on the AlpacaEval2.0 dataset. (a) shows the TTFT distributions, (b) presents the SLO violation rates under varying request-arrival rates,  and (c) compares the end-to-end latency (covering the entire execution from the reasoning phase to the answering phase) across multiple latency metrics under high request-arrival rates.}
    \label{fig:adaptive}
\end{figure}
}

{\bf Effectiveness of adaptive migration.} 
To evaluate the effectiveness of \mbox{\proposed}'s adaptive migration policy, we compare our \mbox{\proposed} against \mbox{\nonadaptive}, which turns off adaptive migration and always triggers request migration at phase transitions, regardless of memory availability. \mbox{\fig{fig:adaptive}}(a) and (b) show the TTFT distributions and SLO violation rates of \mbox{\nonadaptive} versus \mbox{\proposed}. Although the TTFT distributions appear similar, the SLO violation rate under \mbox{\nonadaptive} rises sharply with increasing request arrival rates, reaching 7.45\% compared to only 0.69\% under \mbox{\proposed}. This degradation occurs because \mbox{\nonadaptive} ignores each instance’s current memory state, migrating requests even when the target instance lacks sufficient GPU memory, while the current instance could have immediately served the answering phase without stalling (see \mbox{\fig{fig:load_adaptive}}(a)). Although the instance-level scheduler excludes instances currently violating their SLOs, its decisions rely solely on the current token pacer state and cannot foresee future memory contention due to unpredictable output lengths. Consequently, such unnecessary migrations increase the risk of future SLO violations for answering requests on the target instance, a problem effectively mitigated by \mbox{\proposed}’s adaptive migration policy. Overall, compared to \mbox{\proposed}, \mbox{\nonadaptive} worsens \mbox{{median}} end-to-end latency by 20.1\% and tail latency by 9.7\% (\mbox{\fig{fig:adaptive}}(c)).

\begin{figure}[t]
    \centering
    \includegraphics[width=0.90\columnwidth]{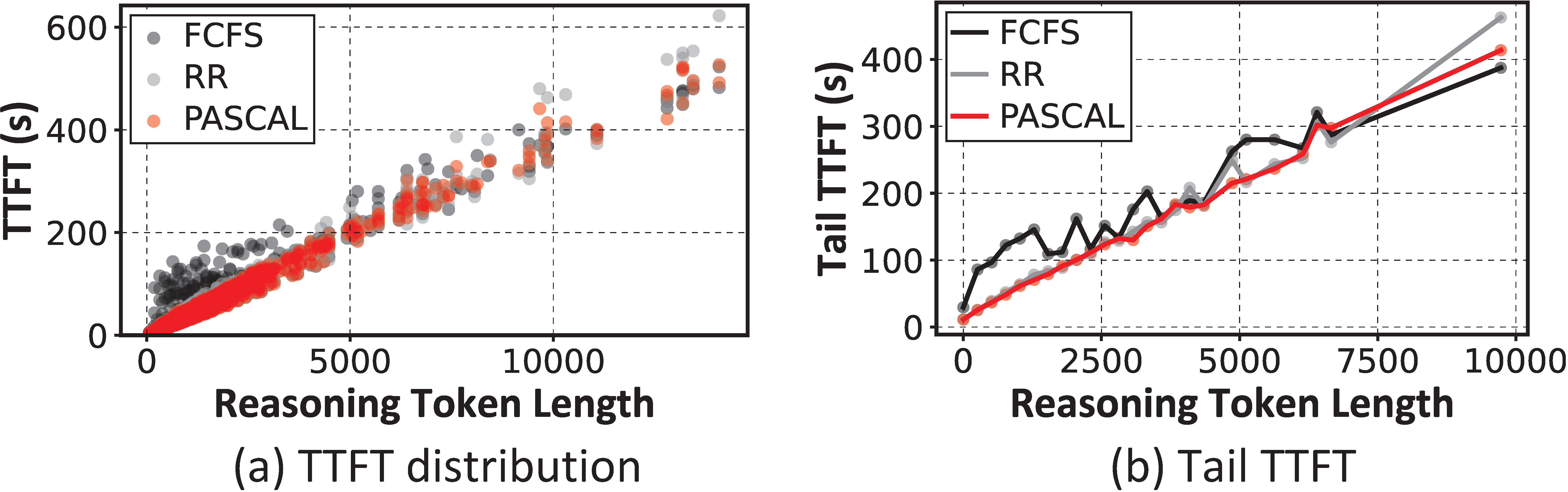}
    \caption{
    {
(a) TTFT distribution and (b) tail TTFT by reasoning token length, with 50\% of the Arena‑Hard trace replaced by the reasoning‑heavy datasets shown in \fig{fig:token_distribution_discussion}. 
}
}
    \label{fig:discussion}
    \vspace{-1.2em}
\end{figure}

{\bf Alternative reasoning datasets.} Our evaluation so far has focused on chat applications where users expect detailed responses with long answering tokens. In contrast, problem‑solving benchmarks~\cite{math-500, gpqa, livecodebench, swe, aime, frames} feature lengthy reasoning phases and comparatively short answers. As shown in \fig{fig:token_distribution_discussion}(a)-(c), the number of reasoning tokens can reach as much as $8.48\times$ higher than answering tokens in these datasets. In such \emph{reasoning‑heavy} datasets, \mbox{\proposed}’s benefits may diminish because answering phases are too short to cause significant scheduling contention, reducing the impact of phase‑aware scheduling.

To this end, we also evaluate a scenario where 50\% of Arena‑Hard requests are replaced with reasoning-heavy requests sampled uniformly from MATH-500, GPQA, and LiveCodeBench. As shown in \fig{fig:discussion}, \proposed reduces tail TTFT by up to 70\% over FCFS for shorter reasoning segments by mitigating head‑of‑line blocking, where long reasoning monopolizes GPU memory until answering completes. Even for long‑reasoning requests, TTFT rises only 6.8\% compared to FCFS, a modest trade‑off given the broader gains. Compared to RR, improvements are smaller due to dataset characteristics: short answering phases create minimal interference with reasoning even under baseline scheduling. RR’s implicit per‑request hierarchy (favoring reasoning over answering, as described in \mbox{\sect{sect:results}}) further reduces contention. Thus, \mbox{\proposed}’s explicit phase isolation provides limited additional benefit in this setting. Nonetheless, \mbox{\proposed} still reduces tail TTFT for certain token ranges by up to 13.9\%, with worst‑case degradation under 7.7\%. SLO violation rates remain similar to RR and consistently lower than FCFS. Overall, \mbox{\proposed} provides stable, competitive performance even when reasoning–answering contention is minimal.

\section{Related works}
\label{sect:related_works}

{\bf LLM serving optimization.}
Prior work~\cite{orca, vllm, aqua, llumnix, podattention, dynamollm, sarathi} has optimized LLM serving for performance and SLO compliance. ORCA~\cite{orca} introduces iteration‑level scheduling with continuous batching to reduce queuing latency and improve throughput. vLLM~\cite{vllm} combines PagedAttention with continuous batching for high‑throughput inference under fine‑grained GPU memory management. Andes~\cite{andes} proposes a QoE‑aware preemptive scheduler for streaming services to preserve user‑perceived responsiveness. Llumnix~\cite{llumnix} supports priority‑aware migration by assigning virtual memory usage to requests, guiding dynamic rescheduling to balance load and prioritize high‑SLO requests. None of these systems address the unique characteristics of the reasoning phase in reasoning LLMs, leaving TTFT optimization largely unexplored.

{\bf Reasoning-based LLM inference.}
Recent work has tackled inefficiencies in long‑chain reasoning. ThinkPrune\mbox{~\cite{hou2025thinkprune}} reduces redundant reasoning steps via reinforcement learning to shorten token sequences without sacrificing performance. 
Dynasor\mbox{~\cite{dynasor}} complements this by predicting requests that are likely to require extended reasoning and allocating additional compute resources accordingly. For requests predicted to have reached a stable or confident reasoning state, Dynasor terminates decoding early to save computation.
In contrast, \mbox{\proposed} is a scheduling framework that exploits the asymmetric resource demands of reasoning versus answering phases, enabling priority‑aware execution and dynamic migration across serving instances without modifying models or adding inference‑time heuristics. Rather than manipulating the reasoning phase itself, \mbox{\proposed} introduces a novel phase-aware scheduling mechanism that distinguishes between the two phases while keeping the execution pipeline intact, making it orthogonal to reasoning-token reduction methods\mbox{~\cite{hou2025thinkprune, dynasor}}.

{\bf Disaggregated LLM inference.}
{
Several prior works~\cite{splitwise,distserve, windserve} leverage the distinct compute characteristics of the prefill and decode stages by disaggregating these stages across dedicated hardware, improving energy efficiency and reducing request contention. DistServe~\cite{distserve}, for example, partitions the inference pipeline based on the observation that prefill and decoding interfere with each other due to their differing resource demands, assigning each stage to hardware configurations that better satisfy their stage‑specific SLO constraints. In contrast to these disaggregated LLM serving systems, which focus on conventional LLMs, our work targets emerging reasoning‑based LLMs, where the decoding stage itself consists of two semantically distinct phases: reasoning and answering. The key contribution of \mbox{\proposed} is identifying this phase‑based execution behavior and intelligently scheduling decoding requests to minimize interference. This insight is orthogonal to existing disaggregated inference systems and can be layered on top of them for additional performance gains.
}

\section{Future work}
\label{sect:future_works}

{\bf \proposed in heterogeneous system architectures.} 
\mbox{\proposed} focuses on GPU-centric LLM serving, consistent with many representative prior works\mbox{~\cite{vllm, splitwise, sarathi, distserve}} where all computations are executed exclusively on GPUs and the CPU is used primarily as a backing storage for offloaded tensors, without participating in the LLM computation process. Meanwhile, several recent studies have explored a \emph{heterogeneous} computing model for LLM inference, in which the CPU collaborates with the GPU to accelerate LLMs\mbox{~\cite{na2024understanding, kim2024exploiting}}.
More recently, the introduction of Rubin CPX\mbox{~\cite{rubin}}—a prefill-optimized GPU designed to complement the standard (now relatively decode-optimized) Rubin GPU\mbox{~\cite{normal_rubin_gpu_not_cpx}} by handling the compute-bound prefill stage—has gained attention as it presents opportunity to balance compute and memory utilization across heterogeneous GPUs (i.e., prefill-optimized vs. decode-optimized).

Applying \mbox{\proposed}'s phase-aware scheduling algorithm on top of these emerging heterogeneous compute devices introduces unique design challenges. For instance, offloading KV caches, activations, or weights must be handled more carefully, as these data may still be needed for ongoing computations depending on how compute and memory operations are partitioned across devices, as well as how scheduling policies and inter-device coherence are orchestrated. A deeper exploration of \mbox{\proposed} in such heterogeneous environments is beyond the scope of this work and is left for future investigation.

{\bf Explicit resource partitioning across reasoning and answering phases.} 
In prior works such as DistServe\mbox{~\cite{distserve}} and Splitwise\mbox{~\cite{splitwise}}, the prefill (compute-bound) and decoding (memory-bound) stages exhibit fundamentally distinct computational characteristics, which makes explicit resource partitioning beneficial. Specifically, DistServe emphasizes the interference between the prefill and decoding stages that stems from their per-step latency difference, where the slower prefill stage delays decoding progress within the same iteration, while Splitwise leverages hardware heterogeneity to better suit each stage’s computational characteristics. 
In contrast, under \mbox{\proposed}, both the reasoning and answering phases belong to the \emph{same} decoding stage. Thus, applying a DistServe-style explicit partitioning offers little benefit, as the two phases have similar per-step latencies and do not experience the iteration-level interference that motivates stage separation. Similarly, applying a Splitwise-style explicit resource partitioning is less effective, since both phases share comparable computational characteristics and are equally well-suited to the same hardware resources, yielding no efficiency gain from assigning them to different devices. Consequently, the potential benefits of explicitly partitioning resources across the reasoning and answering phases is questionable, while its drawbacks become more pronounced. A deeper exploration of this design space is left for future work.

\section{Conclusion}
\label{sect:conclusion}

In this paper, we introduce a phase‑aware scheduling algorithm for reasoning‑based LLM serving systems. Based on our key observation that reasoning latency dominates Time‑To‑First‑Token (TTFT), \proposed's scheduler applies lightweight phase detection and priority assignment during decoding, enabling our proposed system to honor the asymmetric performance sensitivities of reasoning versus answering tokens. A hierarchical intra‑instance queue, combined with SLO‑aware inter‑instance migration, minimizes tail TTFT while preserving answer‑phase token throughput. Consequently, \proposed delivers consistently low tail latency without sacrificing overall throughput or answer quality. 

\section*{Acknowledgment}
\label{sect:ack}
{This work was partly supported by Institute of Information \& Communications Technology Planning \& Evaluation(IITP) grant funded by the Korea government(MSIT) (No.RS-2024-00438851, (SW Starlab) High-performance Privacy-preserving Machine Learning System and System Software), (No.RS-2025-02264029, Implementation and Validation of an AI Semiconductor-Based Data Center Composable Cluster Infrastructure, 30\%), (No.RS-2025-02214652, Development of SoC Technology for AI Semiconductor-Converged Pooled Storage/Memory), (No. RS-2024-00457882, AI Research Hub Project), Samsung Research Funding Center of Samsung Electronics (SRFC-IT2402-03), and IITP under the Graduate School of Artificial Intelligence Semiconductor(IITP-2026-RS-2023-00256472) grant funded by the Korea government(MSIT). We also thank Yunjae Lee for his constructive feedback that helped improve this paper. Minsoo Rhu is the corresponding author.}

\bibliographystyle{IEEEtranS}
\bibliography{refs}

\end{document}